\pdfoutput=1

\documentclass[11pt]{article}

\usepackage[]{acl}

\usepackage{times}
\usepackage{latexsym}

\usepackage[T1]{fontenc}

\usepackage[utf8]{inputenc}

\usepackage{microtype}

\usepackage{listings}
\usepackage{amsmath}
\usepackage{soul} 
\usepackage[skins]{tcolorbox} 
\usepackage{graphicx} 
\usepackage{multirow} 
\usepackage{booktabs} 
\usepackage{multirow} 
\usepackage{array,booktabs,ragged2e} 
\newcolumntype{R}[1]{>{\RaggedLeft\arraybackslash}p{#1}} 
\newcolumntype{L}[1]{>{\RaggedRight\arraybackslash}p{#1}} 
\usepackage{subcaption} 
\newcommand{\hide}[1]{}

\usepackage{calc}
\newlength\myheight
\newlength\mydepth
\settototalheight\myheight{Xygp}
\newcommand{\best}[1]{\colorbox{black!10}{#1}}
\newtcbox{\pmarker}{enhanced,nobeforeafter,tcbox raise base,boxrule=0.4pt,top=0mm,bottom=0mm,
  right=0mm,left=0mm,arc=1pt,boxsep=2pt,before upper={\vphantom{dlg}},
  colframe=purple!50!black,coltext=purple!25!black,colback=purple!10!white}

\newtcbox{\marker}{enhanced,nobeforeafter,tcbox raise base,boxrule=0.4pt,top=0mm,bottom=0mm,
  right=0mm,left=0mm,arc=1pt,boxsep=2pt,before upper={\vphantom{dlg}},
  colframe=red!50!black,coltext=red!25!black,colback=red!10!white}
  
\newtcbox{\greymarker}{enhanced,nobeforeafter,tcbox raise base,boxrule=0.4pt,top=0mm,bottom=0mm,
  right=0mm,left=0mm,arc=1pt,boxsep=2pt,before upper={\vphantom{dlg}},
  colframe=black!50!white,coltext=black,colback=black!10!white}

\newtcbox{\one}{enhanced,nobeforeafter,tcbox raise base,boxrule=0.4pt,top=0mm,bottom=0mm,
  right=4mm,left=0mm,arc=1pt,boxsep=2pt,before upper={\vphantom{dlg}},
  colframe=blue!50!black,coltext=blue!25!black,colback=black!10!white,
  overlay={\begin{tcbclipinterior}\fill[blue!75!white] (frame.south east)
    rectangle node[text=white,font=\sffamily\bfseries\small] {1} ([xshift=-4mm]frame.north east);\end{tcbclipinterior}}}
    
\newtcbox{\two}{enhanced,nobeforeafter,tcbox raise base,boxrule=0.4pt,top=0mm,bottom=0mm,
  right=4mm,left=0mm,arc=1pt,boxsep=2pt,before upper={\vphantom{dlg}},
  colframe=blue!50!black,coltext=blue!25!black,colback=black!10!white,
  overlay={\begin{tcbclipinterior}\fill[blue!75!white] (frame.south east)
    rectangle node[text=white,font=\sffamily\bfseries\small] {2} ([xshift=-4mm]frame.north east);\end{tcbclipinterior}}}

\newcommand\blfootnote[1]{%
  \begingroup
  \renewcommand\thefootnote{}\footnote{#1}%
  \addtocounter{footnote}{-1}%
  \endgroup
}

\newcommand*{\img}[1]{%
    \raisebox{-.1\baselineskip}{%
        \includegraphics[
        height=\baselineskip,
        width=\baselineskip,
        keepaspectratio,
        ]{#1}%
    }%
}

\newcommand*{\imgg}[1]{%
    \raisebox{-.3\baselineskip}{%
        \includegraphics[
        height=\baselineskip,
        width=\baselineskip,
        keepaspectratio,
        ]{#1}%
    }%
}

\definecolor{lpink}{cmyk}{0, 0.7808, 0.4429, 0.1412}
\definecolor{aqua}{cmyk}{0.91, 0, 0.09, 0.36}
\definecolor{ao}{rgb}{0.0, 0.5, 0.0}
\definecolor{amber}{rgb}{1.0, 0.49, 0.0}
\definecolor{dblue}{rgb}{0.0, 0.0, 0.61}
\definecolor{burgundy}{rgb}{0.5, 0.0, 0.13}

\definecolor{bnk}{RGB}{50, 50, 100}
\newenvironment{belief}{\fontfamily{FPL Neu}\selectfont}{\par}%
\DeclareTextFontCommand{\textbelief}{\belief}
\newcommand{\data}{{\textsc{{NormBank}}}}
\newcommand{\datae}{\textsc{NormBank}}
\newcommand{\framework}{SCENE}

\newcommand{\setting}{\imgg{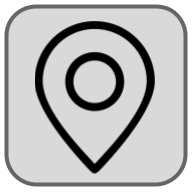}}
\newcommand{\behavior}{\imgg{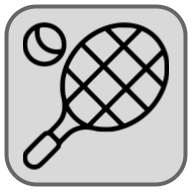}}
\newcommand{\role}{\imgg{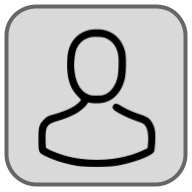}}
\newcommand{\environment}{\imgg{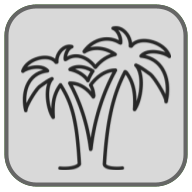}}
\newcommand{\attribute}{\imgg{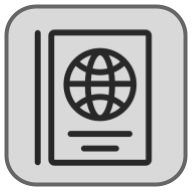}}

\definecolor{unexpected}{RGB}{152, 0, 0}
\definecolor{okay}{RGB}{74, 134, 232}
\definecolor{expected}{RGB}{56, 118, 29}

\newcommand{\numAnnotatedNorms}{155k}
\newcommand{\numAnnotatedConstraints}{408k} 
\newcommand{\numUniqueConstraints}{63k} 
\newcommand{\numUniqueConjunctions}{70k} 
\newcommand{\constraintsPerConjunction}{2.63} 
\newcommand{\numAttributeConstraints}{$578$}
\newcommand{\numEnvironmentConstraints}{$404$}

\newcommand{\taxonomyCoverage}{94\%}
\newcommand{\taxonomyValueCoverage}{69\%}

\newcommand{\numStagingHITs}{$2,502$}

\title{\datae{:} A Knowledge Bank of Situational Social Norms}

\newcommand{\cqa}[1]{#1{\color[HTML]{CB4335}$\dagger$}}
\newcommand{\siqa}[1]{#1{\color[HTML]{2E86C1}$\ddagger$}}
\newcommand{\base}[1]{#1{$^\star$}}

\newcommand{\treelogo}{\raisebox{5pt}{\includegraphics[scale=0.050]{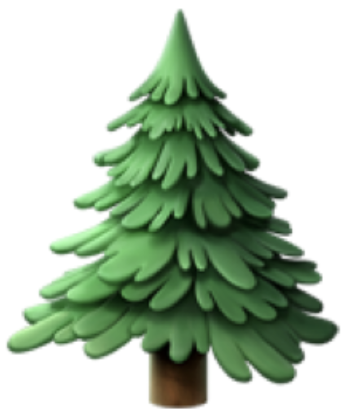}}}
\newcommand{\metalogo}{\raisebox{4pt}{\includegraphics[scale=0.004]{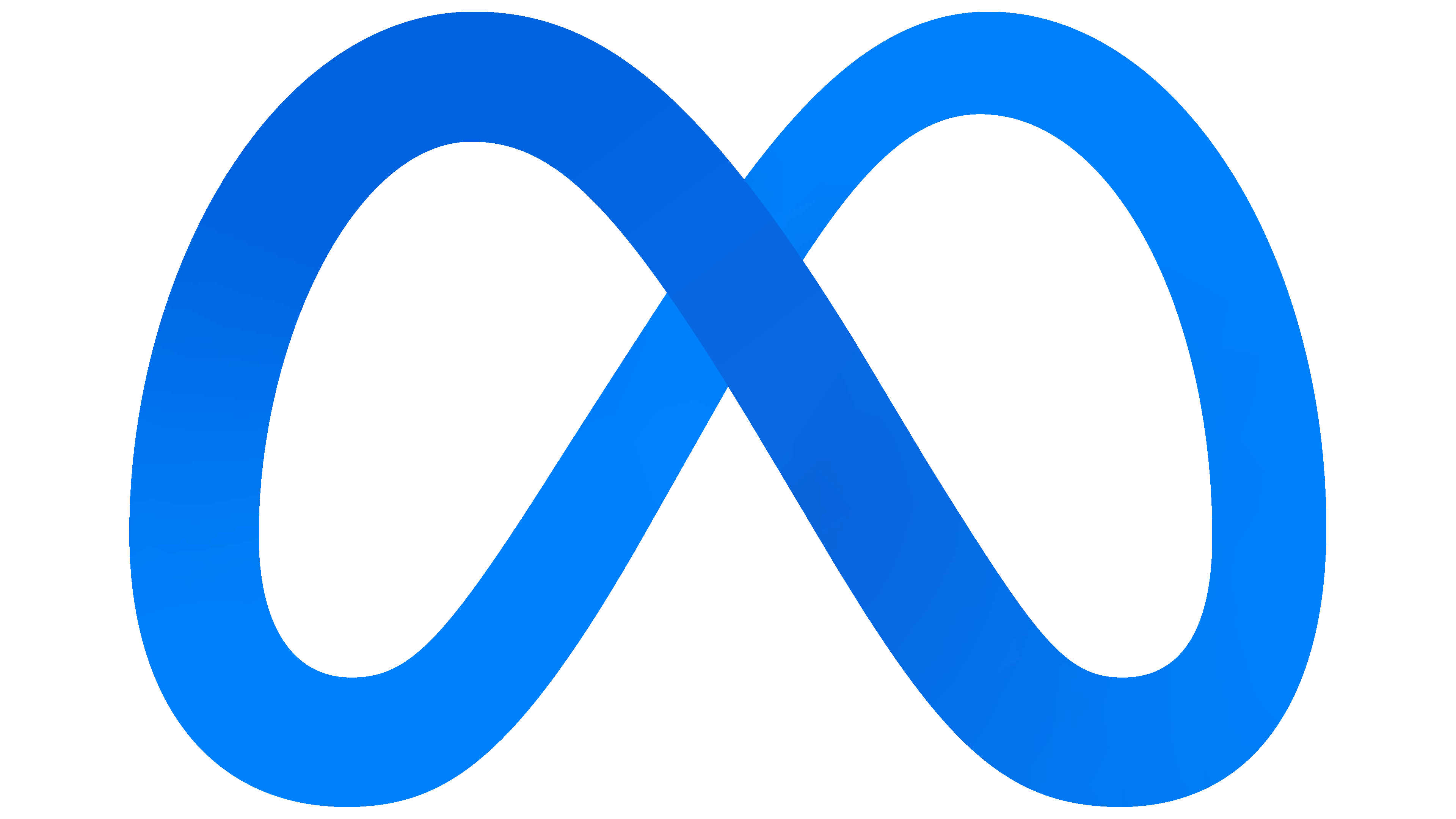}}}

\newcommand{\sunglasses}{\raisebox{3pt}{\includegraphics[scale=0.08]{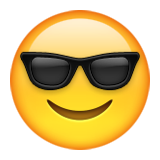}}}
\newcommand{\stanf}{\treelogo}
\newcommand{\meta}{\metalogo}
\newcommand{\intern}{\sunglasses}

\author{Caleb Ziems \stanf \intern \hspace{0.3em}
        Jane Dwivedi-Yu \meta \hspace{0.3em}
        Yi-Chia Wang \meta \hspace{0.3em}\\
        \textbf{Alon Y. Halevy} \meta \hspace{0.3em}
        \textbf{Diyi Yang} \stanf \\
        \stanf Stanford University, \meta Meta AI\\
        \texttt{\small \{cziems, diyiy\}@stanford.edu},
        \texttt{\small \{janeyu, yichiaw, ayh\}@fb.com}
}

\begin{document}
\maketitle
\begin{abstract}
We present \data{}, a knowledge bank of \numAnnotatedNorms{} situational norms. This resource is designed to ground flexible normative reasoning for interactive, assistive, and collaborative AI systems. Unlike prior commonsense resources, \data{} grounds each inference within a multivalent sociocultural {frame}, which includes the setting (e.g., \textit{restaurant}), the agents' contingent roles (\textit{waiter}, \textit{customer}), their attributes (\textit{age}, \textit{gender}), and other physical, social, and cultural constraints (e.g., the \textit{temperature} or the \textit{country of operation}). In total, \data{} contains \numUniqueConstraints{} unique constraints from a taxonomy that we introduce and iteratively refine here. Constraints then apply in different combinations to frame social norms. Under these manipulations, norms are \textit{non-monotonic} --- one can cancel an inference by updating its frame even slightly. Still, we find evidence that neural models can help reliably extend the scope and coverage of \data{}. We further demonstrate the utility of this resource with a series of transfer experiments. For data and code, see {\small \url{https://github.com/SALT-NLP/normbank}}
\blfootnote{\intern Work done at Meta AI Research}
\end{abstract}

\section{Introduction}
\label{sec:introduction}

\begin{figure}
    \centering
    \includegraphics[width=0.852\columnwidth]{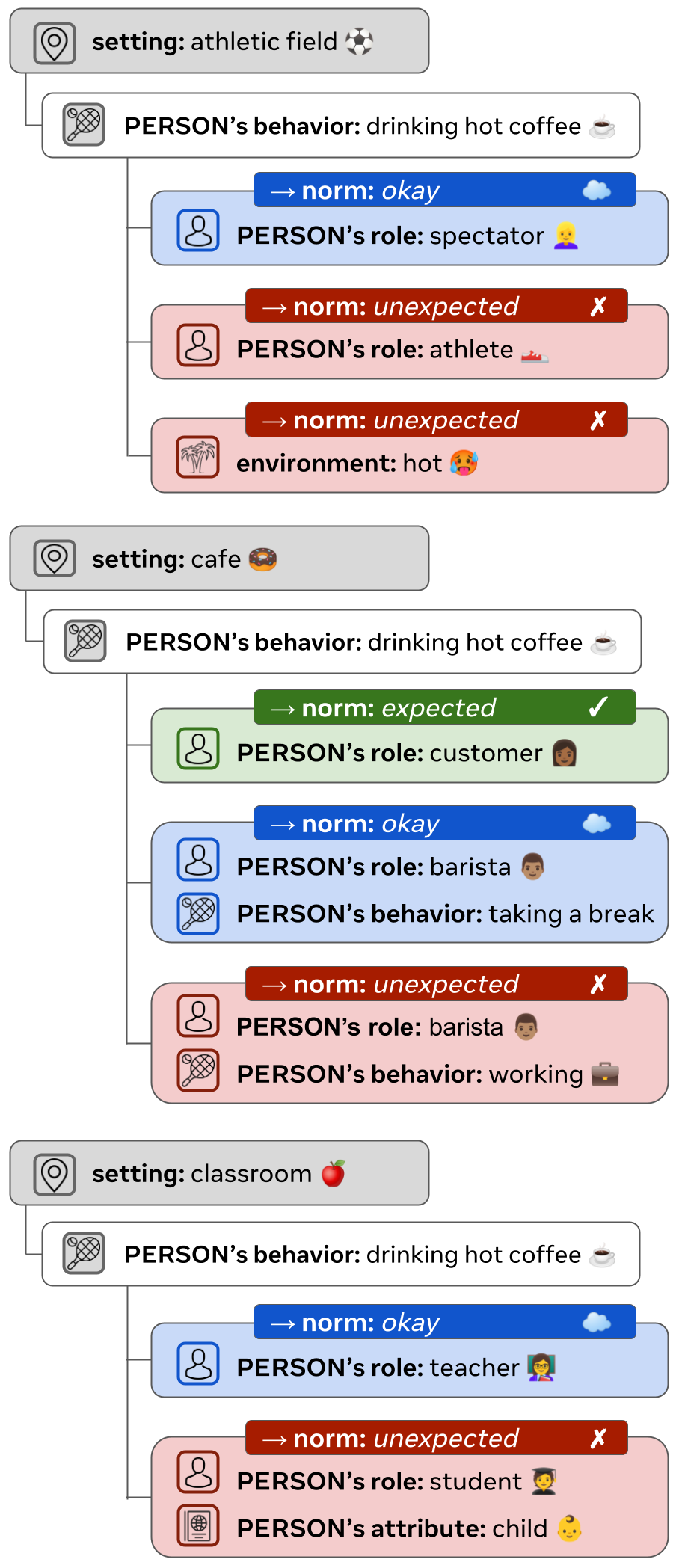}
    \caption{\textbf{What is special about \datae{}?} Norms are grounded by \textit{situational constraints}---environmental and personal attributes, as well as roles and other behaviors. In this example, drinking coffee is an encouraged activity in its prototypical context, for a \textit{customer} in a \textit{cafe}, but it is counternormative for a working barista to do so in the same cafe, or for a child-age student do so in a classroom. These represent only some of the non-monotonic normative inferences that are represented in \data{}.}
    \label{fig:crown_jewel}
\end{figure}

As AI systems continually evolve for human assistance and collaboration, they will increasingly operate within cultural and social spaces, and require increasingly robust and flexible knowledge of social norms \citep{carlucci2015explicit}. From dialogue systems \citep{molnar2018role,vaidyam2019chatbots,bavaresco2020conversational,grossman2019mathbot} to socially interactive robots \citep{fong2003survey,deng2019embodiment} and augmented or mixed reality technologies \citep{anderson2022metaverse}, each could benefit from understanding how humans effectively communicate, make decisions, engage with requests, and broadly interact with others \citep{sunstein1996social,sherif1953groups}.

Natural language is flexible and highly expressive; thus it is a promising medium for encoding knowledge of social norms \citep{sap2019atomic}. The goal of this work is to construct \data{}, a natural language bank of social norms that will allow AI systems to reason about social situations under complex constraints. \data{} encodes \numAnnotatedNorms{} norms via scalable human annotation, bootstrapped with implicit knowledge from large pre-trained language models (LLMs).

\data{} factors in two important considerations that have been previously overlooked. First: norms are not rigid truths; they are flexibly assumed standards that may be updated with new information from the social context \citep{blass2015implementing}. Second: this social context is not a flat list of facts but a matrix of hierarchical dependencies \citep{hovy2021importance}. These two considerations have important design implications for norm representations and reasoning in AI, inspiring two objectives for this work.

\textbf{Objective 1} is to support \textit{non-monotonic} reasoning \citep{reiter1981closed} over \textit{defeasible} \citep{pollock1987defeasible} norms. This means inferences that hold under most cases can be updated or even retracted based on new information. For example, \textit{dancing} is a positive behavior that is generally permitted in many casual settings and in many cultures.
We can still strengthen or cancel this inference. On the one hand, dancing is expected from a professional dancer. But in an Islamic cultural context, individuals are forbidden from publicly dancing with members of the opposite sex. In a hospital setting, a young child is allowed to dance in the waiting room, but this behavior would not be expected from an adult visiting a dying relative. For more examples, see Figure~\ref{fig:crown_jewel}. This kind of reasoning will {not} always follow straightforward compositional logic \citep{klimczyk2021compositional}, and we expect it to be a challenge for AI systems.\footnote{Typically, \textbelief{it's okay to drink soda while actively working} and \textbelief{it's okay for a waiter to drink soda}; yet the intersection of these conditions is not typical. \textbelief{It is } \texttt{\textbf{NOT}} \textbelief{okay for a waiter to drink soda while actively working.}} \data{} is the first data resource to support non-monotonic normative reasoning by encoding \textit{contrasting} situations under which the \textit{same behavior} could alternatively be expected or considered taboo (see \S\ref{sec:annotation}).

There is a combinatorially explosive space of situational contexts, each with non-compositional and thus unpredictable norms. Enumerating the set of all possible constraints is intractable. To efficiently learn norms in this space, models can rely on the regularizing effects of hierarchical organization and social theory. \data{} introduces hierarchical organization (\textbf{Objective 2}) by means of a rich taxonomy over the relevant contextual signals that inform behaviors.

Our new \framework{} taxonomy is the first to use \citeauthor{goffman1959presentation}'s (\citeyear{goffman1959presentation}) dramaturgical theory of social life. We operationalize the theory with \textit{settings} that have additional \textit{environmental} constraints. In each setting, there are agents with different \textit{roles} and \textit{attributes}, who then perform \textit{behaviors}. Norms apply to behaviors in certain situations. For example, in Figure~\ref{fig:crown_jewel}, norms around \textit{drinking hot coffee} differ for agents with different roles (e.g., \textit{barista, customer}) and attributes (e.g., \textit{adult, child}), in different settings (e.g., \textit{cafe, classroom}).

Having addressed the objectives above, we train neural models to expand \data{} with automatic knowledge completion. Experiments show promising results: these models can extrapolate social commonsense to new behaviors in new situations, leveraging similarities in analogous roles across different situations. Finally, we demonstrate how to transfer knowledge via sequential finetuning from \data{} to social reasoning tasks. Together, knowledge completion and transfer learning suggest that our dataset will serve as a useful resource for advancing neural models toward situationally-grounded social reasoning.

\section{Related Work}
\label{sec:related_work}

\paragraph{Commonsense knowledge bases} (CSKBs) are sets of structured knowledge about everyday life. They capture broad taxonomic relationships \citep{liu2004conceptnet, speer2017conceptnet,elsahar2018t}, logical relations \citep{lenat1995cyc,zhang2018record}, and universal laws of causality and physical mechanics \citep{talmor2019commonsenseqa,bisk2020piqa}. More recent datasets encode \textit{social mechanics}, like broad human values \citep{ziems2022mic, hendrycks2021aligning}, 
norms \citep{forbes2020social,fung2022normsage}, and typical rules of social behavior and motivation \citep{sap2019atomic,huang2019cosmos}. 
NormBank specifically centers social norms around dramaturgical settings (i.e., \textsl{places of worship, commerce, and recreation}).
Just as ATOMIC \citep{sap2019atomic} seeded Social IQa \citep{sap2019social} and $\delta$-NLI \citep{rudinger2020thinking}, we anticipate that NormBank can be converted into benchmarking tasks, plus injected into language models for downstream applications \citep{chang-etal-2020-incorporating,mitra2019additional,ji2020generating,ji2020language}.

\paragraph{Norm discovery} is an emerging method inspired by automatic knowledge base construction \citep{mitchell2018never,weston2013connecting,craven2000learning} and extracting social knowledge from LLMs via prompting \citep{trinh2018simple,petroni-etal-2019-language,wang-etal-2019-make,sakaguchi2020winogrande}. In concurrent work, \citet{fung2022normsage} propose \textsc{NormSAGE}, which automatically discovers {mandated} or {conventional} behaviors from dialogues. Their prompts resemble our bootstrapping efforts in \S\ref{sec:scene_framework},
with the added step of automatic self-verification. \data{} differs from \textsc{NormSAGE} in that we rely on human annotation to collect more creatively non-prototypical situations to challenge and expand normative reasoning models.

\paragraph{Normative reasoning} systems like Delphi \citep{jiang2021delphi}, and UNICORN \citep{lourie2021unicorn} are pre-trained on existing social knowledge bases \citep{forbes2020social, emelin2021moral, hendrycks2021aligning, lourie2021scruples, sap2020social}, which contain more conventional social behaviors from narrative contexts. Until \citet{pyatkin2022reinforced}, in work concurrent to our own, descriptive social reasoning systems have been framed as universal oracles with forced-choice judgments about human behaviors \citep{talat-etal-2022-machine}. These models lack the capacity for defeasible reasoning \citep{madaan2021could,rudinger2020thinking}. Oracles instead tend to assume the most prototypical contexts \citep{boratko2020protoqa}. Many of these predictions will appear reasonable if we pragmatically infer a conventional narrative, but for systems to achieve robust social intelligence, they must account for the long tail of the distribution. We can easily find unconventional contexts in which the correct inference contained in \data{} is misunderstood by current models.\footnote{For example, Delphi believes \textit{yelling} and \textit{clenching your fists}, \textit{breathing heavily}, or \textit{asking someone personal questions about their sex life} are all conventionally inappropriate. \data{} gives acceptable contexts for each: \textit{guests riding a roller coaster, athletes running track}, \textit{and doctors performing routine checkups,} respectively.}

\begin{figure*}
    \centering
    \includegraphics[width=1\textwidth]{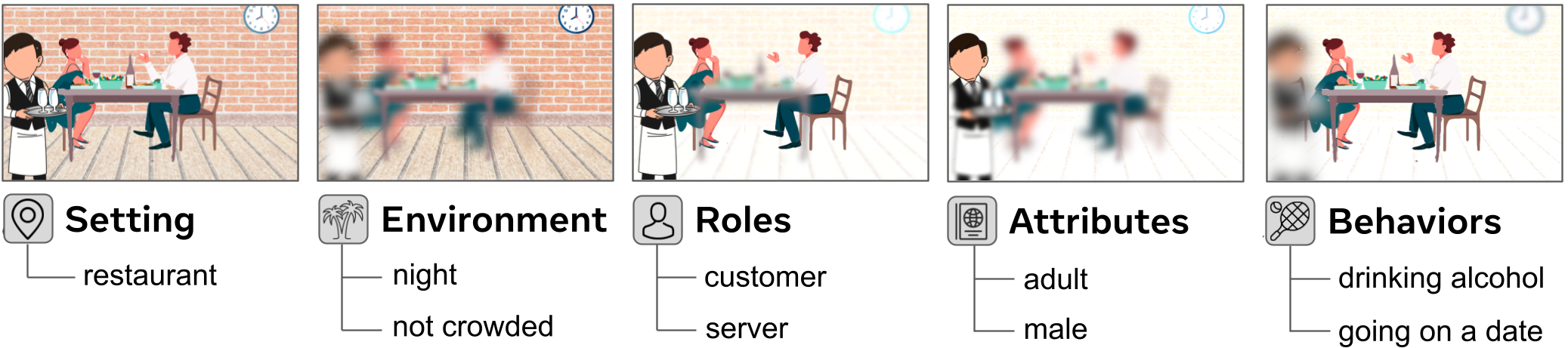}
    \caption{\textbf{An example of the \framework{} Dramaturgical Framework used to constrain \data{}}. The \textit{restaurant} setting is specified by the \textbf{attendance} (\textit{not crowded}) and \textbf{time of day} (\textit{night}) in the environment. The two agent roles, \textit{customer and server}; the latter is specified by the \textbf{age bracket} (\textit{adult}) and \textbf{gender} (\textit{male}) attributes. The former are engaged in the behaviors \textit{drinking alcohol} and \textit{going on a date.} \textit{Note:} Graphics are for illustration. \data{} is a text dataset and does not contain any images.}
    \label{fig:scene_elements}
\end{figure*}

\section{SCENE: A Dramaturgical Framework}
\label{sec:scene_framework}
\begin{quote}
    {
    \textsl{The self... is a dramatic effect arising diffusely from a SCENE.} \phantom{a} \img{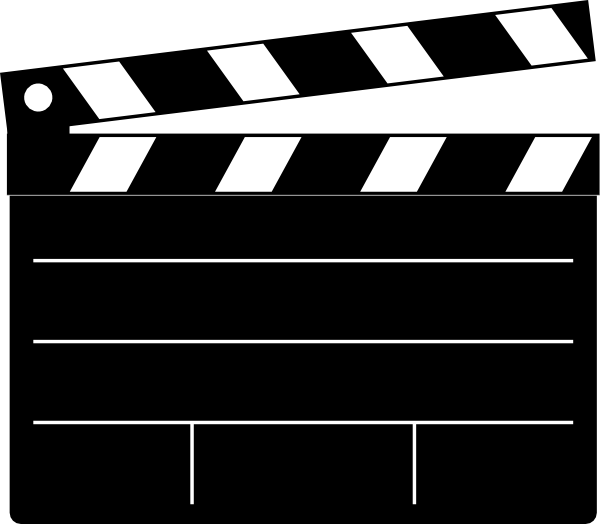} \\\phantom{abcdefg}--- \textbf{Erving Goffman} (\citeyear{goffman1959presentation})
    }
\end{quote}

To help models efficiently learn non-monotonic normative reasoning over a seemingly unbounded set of possible contexts, and to test this understanding in LLMs, we will need to establish a more tractable set of elements to represent this social matrix. For this purpose, we construct a hierarchical taxonomy of constraints, which we call \textit{\underline{S}ituational \underline{C}onstraints for social \underline{E}xpectations, \underline{N}orms, and \underline{E}tiquette} (\framework{} for short). SCENE follows \citeauthor{goffman1959presentation}'s (\citeyear{goffman1959presentation}) dramaturgical model of social life. According to this model, people are like actors trying to maintain a social {performance} in front of an audience. Each actor performs a particular \textit{role} as if in a scene from a movie. The scene is grounded in a particular \textit{setting}, which includes aspects of the \textit{environment} that inform the performance. Each scene also has a script \citep{schank1977scripts}, which tells the actor what kinds of \textit{behaviors} will be perceived as in-character or out-of-character. Additionally, the actor will embody socially meaningful \textit{attributes} like age, gender, status, etc. These attributes may be relevant to the scene and the actors place in it. In Figure~\ref{fig:scene_elements}, the example setting is a \textit{restaurant} where the environment is \textit{uncrowded} and the hour is \textit{night}. There are two primary roles of \textit{customer} and \textit{server}, and for norm formation, some relevant attributes include their respective \textit{genders}, \textit{sexualities}, and \textit{ages}, which parameterize the behaviors that are appropriate for this dinner, such as \textit{dating} and \textit{drinking alcohol}.

\paragraph{Settings \setting{}} 
(e.g., banks, classrooms, homes, hospitals) are the loci of scripted social interactions \citep{schank1977scripts}, and they frame all subsequent elements of \data{}, so we begin here with 129 distinct settings like \textit{amusement park, bus,} and \textit{elevator}. Settings derive from two popular knowledge resources. First, there are 80 settings from from ConceptNet \citep{speer2017conceptnet}, a broad knowledge base of the words and phrases that people commonly use.\footnote{Settings were defined by head-entities with an \texttt{IsA} relationship to some tail in {\small\{\texttt{place}, \texttt{location}, \texttt{area}\}}. Manual inspection proved the usefulness of this heuristic.} There are another 255 settings from the ``movie scene'' label in the MovieGraphs \citep{vicol2018moviegraphs} resource---a collection of social situations that were depicted in movie clips.

\paragraph{The Environment \environment{}}
contains signals that can trigger associative priming of social norms (e.g., the noise level of a study space; \citeauthor{aarts2003silence}). This portion of the taxonomy is designed to be broad and general-purpose, with a set of attributes that can refine any setting. Our taxonomy is based on a broad review of the literature on norm formation and its relevant factors \citep{van2018illuminating,janicik2003talking,boyce2000perceptions,russell1982environmental,durkheim1915elementary}. Importantly, the taxonomy is further refined through crowdsourced feedback (\S\ref{sec:annotation}). Ultimately, our taxonomy grows to contain \numEnvironmentConstraints{} environmental constraints. An extensive overview of the environmental constraints is given in Appendix~\ref{appdx:environment}, but we summarize them here.

In the environment, there are important taxonomic subclasses of factors that inform norms. One subclass is \textbf{time} constraints, like seasonality \citep{janicik2003talking}, holidays and special customary observances \citep{durkheim1915elementary}, and another is the \textbf{country of operation}, which serves as a proxy for regional cultural differences \citep{meyer2014culture}. We also include factors from \textbf{environmental psychology} \citep{bell2001environmental} that involve the agent's comfort and ease in the environment (e.g., noise level, privacy, and cleanliness). Additionally, \textbf{physical conditions} include factors like weather, which impact visibility, coordination, safety, and comfort \citep{boyce2000perceptions,van2018illuminating,cunningham1979weather}. In addition to the imposed taxonomy, annotator feedback (\S\ref{sec:annotation}) lead us to add a subclass called \textbf{restrictions} that formally limit attendance, participation, and behavior, due to notions of formality, religiosity, or exclusiveness.

\paragraph{Roles \role{}}
\label{subsec:roles}
may be ubiquitous, but it is challenging to collect reliable, setting-specific roles with high coverage. Our solution is to use the powerful associative knowledge of LLMs to automatically enumerate roles for each setting via prompting, inspired by \citet{trinh2018simple}, \citet{petroni-etal-2019-language}, and others \citep{wang-etal-2019-make,sakaguchi2020winogrande}. Specifically, we prompt GPT-3 \citep{brown2020language} \texttt{\small text-davinci-002} in a zero-shot manner with the phrase {\small
    ``\texttt{Some roles <preposition> <determiner> <setting>:}''
    }
where the preposition and determiner are manually configured to match the setting (for example, ``\texttt{\small some roles at a casino:}'' or ``\texttt{\small some roles on the beach:}''). On average, we generate 5.5 roles per setting, with a total of 928 unique roles. 

\paragraph{Attributes \attribute{}}
are properties of individual agents that determine their social norms. Here again, the goal is to derive a general purpose taxonomy from the literature. Some attributes are basic demographic categories like the person's \textbf{age bracket}, \textbf{gender}, \textbf{race}, \textbf{religion}, and \textbf{sexuality} \citep{thompson1986structure,dempsey2004cohabits,helgeson2016psychology}. Related demographic categories include \textbf{education} level, \textbf{employment}, and \textbf{marital status}.
Since food is a focal point for culture and morality, we include \textbf{diet}. We also include material constraints like \textbf{medical condition} and \textbf{social class}. Finally, we increase the coverage of this set by including generic descriptors of two types: \textbf{condition or state} adjectives, which describe a temporary condition (e.g., \textit{dizzy}), and \textbf{characteristic} adjectives that describe more permanent attributes (e.g., \textit{blonde}). In total, our taxonomy defines \numAttributeConstraints{} attribute constraints.

\paragraph{Behaviors \behavior{}}
are the primary target of analysis for social norms. As with roles, we co-opt GPT-3 to enumerate behaviors for each setting and role, but the approach here is augmented in two ways. First, we include a norm \textit{expectation} in the prompt. By querying for \textit{unexpected} behaviors, we can begin to shift the distribution of behaviors away from the prototypical. Second, we increase the diversity of generations by conditioning on the agent's attribute. This further reduces the number of conventional behaviors in our set. The prompt is {\small
    ``\texttt{Some things you would (never) do <preposition> <determiner> <setting> (if you were <attribute>):}''}
where elements in parentheses are optional elements. In this way, we generated an average of 776.5 behaviors per setting, which was filtered down to 112.6 behaviors per setting, via programmatic methods described in Appendix~\ref{appdx:behaviors}.

\section{Building \datae{}}
\label{sec:annotation}
Section~\ref{sec:scene_framework} gave us a high-recall set of constraint variables for explaining situational social norms. Our end goal is to build a resource that contains reliable norms to ground, train, and test automatic normative reasoning systems. We want these norms to describe challenging, non-prototypical examples, and to depend on subtly contrasting situations that, when shifted, change the norm label non-monotonically. This motivates us to use human annotation over the rich \framework{} taxonomy. 

Our process is essentially the reverse of the current paradigm established by prior work, which starts with a basic narrative context and subsequently extracts \citep{fung2022normsage} or annotates \citep{forbes2020social} the expected behaviors. Instead, we start with behaviors and ask annotators to provide us with different dramaturgical contexts (\framework{} constraints) under which that behavior could be variously seen as \textit{expected}, \textit{okay}, or \textit{unexpected}. Thus we obtain richer and less prototypical instances---examples not mentioned in standard dialogue, which will significantly challenge models. The approach is inspired by contrast sets \citep{gardner2020evaluating} and counterfactual augmentation \citep{kaushik2019learning} as means of reducing spurious correlations in model inferences.

\begin{table*}
    \centering
        \resizebox{\linewidth}{!}{%
        \begingroup
        \renewcommand{\arraystretch}{1.25} 
        \begin{tabular}{r|r|r|r|r|r|r}
            \toprule
& {Constraints} & {Norms} & {Situations} & {Constr. / Norm} & {Taxonomic Constr.} & {Pre-populated Constr.}\\ \midrule
\texttt{train} & 328,045 & 124,920 & 57,417 & 2.63 & 93.5\% & 69.3\%\\
\texttt{dev} & 37,761 & 15,008 & 8,573 & 2.52 & 92.8\% & 66.8\%\\
\texttt{test} & 42,601 & 15,495 & 8,674 & 2.75 & 94.6\% & 70.9\%\\\hline 
\data{} & 408,407 & 155,423 & 70,215 & 2.63 & 93.6\% & 69.2\%\\

             \bottomrule
        \end{tabular}
        \endgroup
        }
        \caption{\textbf{Summary statistics} show the immense scale of \data{} (\S\ref{sec:annotation}) and the broad coverage of our \framework{} framework (\S\ref{sec:scene_framework}). There are \numAnnotatedNorms{} total annotated norms, comprised of \numUniqueConjunctions{} unique situations, and each situation is drawn from a conjunction of some subset of the \numAnnotatedConstraints{} annotated constraints. Of these annotated constraints, \taxonomyCoverage{} of them use the structure of our \framework{} taxonomy, and \taxonomyValueCoverage{} use a pre-populated constraint value from one of our taxonomic dropdown menus.
        }
        \label{tab:dataset_statistics}
\end{table*}

\subsection{Annotation Task}

For the annotation task, we recruit experienced English-speaking Mechanical Turk annotators who have $\geq$98\% acceptance with $\geq$100 HITs and are located in the United States. The task requires human creativity over a large combinatorial space. For a given setting $s$ and a behavior $b$, an annotator will tell us distinct situational contexts under which $b$ is alternatively \textit{expected} (required by duty or anticipated with high probability), \textit{okay} (permitted or anticipated with moderate probability), or \textit{unexpected} (forbidden, stigmatized, taboo, or otherwise anticipated with very low probability). 

These \textit{expected, okay,} and \textit{unexpected} categories are called ``norm labels.'' The language of expectation is useful for describing behavioral regularities---the focus of this work---rather than enumerating top-down or bottom-up judgements of \textit{ethical} or \textit{moral} behavior, as in prior datasets \citep{ziems2022mic,emelin2021moral,lourie2021scruples}. Importantly, we do not impose any ethical philosophy or framework as in \citet{hendrycks2021aligning}, but instead, encourage annotators to find norms that merely describe observable social life \citep{cialdini1991focus}.

The annotator fully specifies the appropriate situational context by means of disjunctions and conjunctions of constraints. For example, ``spit at a dentist's office'' can be unexpected when \texttt{\small (PERSON's role is `dentist')} or when \texttt{\small{((PERSON's role is `patient') AND (PERSON's behavior is `checking in'))}}. Annotators select \framework{} constraints using drop down menus that follow the hierarchy of \S\ref{sec:scene_framework} (for details on the HIT interface, see Appendix~\ref{appdx:hit_interface}). They are also free to insert their own custom constraints into the hierarchy. In this way, we iteratively expand the taxonomy.

\subsection{Dataset Quality}
\label{subsec:quality_control}
\paragraph{Quality Control.} Manual inspection of over 2.5k data points reveals that the open-ended and creative aspects of the task are natural incentives for high-quality work \citep{chandler2015using,sheehan2018crowdsourcing}.
To further ensure the quality of \data{}, we trained annotators with careful instruction, a qualification test, a staging round, personalized feedback, programmatic filtering, and finally, a series of random audits \citep{litman2015relationship,sheehan2018crowdsourcing}. The instructions included at least 3 fully-worked examples for each norm label, plus suggestions and explanations for a total of 24 constraints. We administered a six-question qualifier, which tested workers' knowledge of the taxonomy, definitions, free text response, and how to properly indicate constraint conjunctions and disjunctions through the task interface. If the worker passed at least five questions correctly on the first try, she would gain access to the staging round -- a small-scale version of the task in which each submission would receive detailed and personalized feedback. 

{We invested a significant amount of time to feedback, offering $75$ to $200$ words of review for each of \numStagingHITs{} staging HITs.} Once a worker submitted 3 high-quality HITs in the staging round, he or she could move to the full task. To identify poor work here, we programmatically flagged workers with extremely low variation in their annotations. Finally, we periodically performed a total of three random audits, sampling $250$ annotations in each audit, to confirm the quality of the annotations. Workers were paid a base rate, plus an additional itemized bonus for every additional constraint they added, which incentivized workers to be more expressive and creative. Annotators received a median of \$30 per hour for this task.

\paragraph{Quality Metrics.}
The above methods all proved remarkably successful in generating a creative and high-quality resource. Because our task is creative and subjective, data quality is not easily measured by inter-annotator agreement. We instead report human evaluations over the Gold \data{} data in the bottom row of Table~\ref{tab:gen} (alongside model generations from \S\ref{subsec:knowledge_completion}). Annotations are considered sensible (82.5\%), relevant (82.17\%), and normative (72.9\%). Still, it is important to note that around half of gold annotations are marked by third-party evaluators as fully correct by majority vote. 

With regard to the correctness metric, annotator disagreements can be traced to differences in the annotators’ models of the world, which likely stem from their own personal differences, including age, profession, and worldview. For example, an annotator likely familiar with the Cambodian tradition of ``Pithi Srang Preah'' marked that ``honoring your ancestors'' is normal for Cambodians on Cambodian New Year, while an annotator unfamiliar with this practice marked it as unexpected. 
Furthermore, we administered political leaning and the moral foundations surveys to all annotators, which we release alongside \data{} to help explain how these personal differences informed the probabilities they assigned to events. This resource will be of interest to computer scientists and social scientists, since \data{} contains not only commonsense facts, but also culturally-conditioned distributions over behavior and expectations about behavior.

\begin{figure}
    \centering
    \includegraphics[width=\columnwidth]{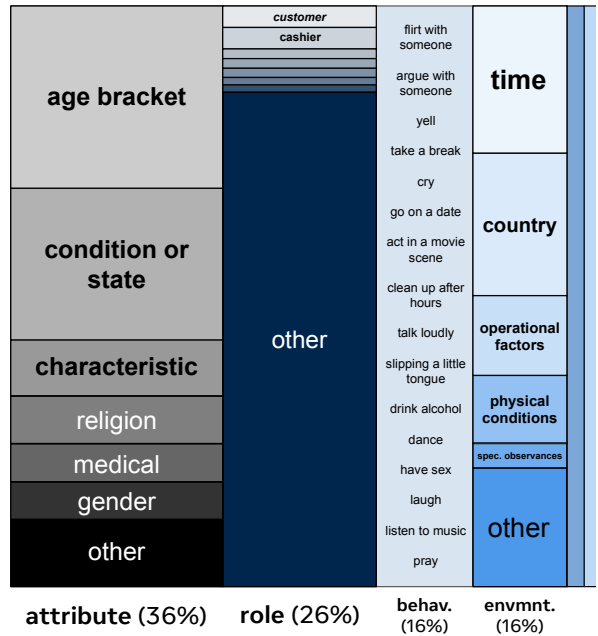}
    \caption{\textbf{Distribution of \data{} constraints} where the area of each cell is proportional to the distribution. Constraints are dominated most by the agent's attributes and roles, with a smaller and even split between behaviors and the environment.}
    \label{fig:dataset_label_proportions}
\end{figure}
\subsection{Dataset Summary}
\label{subsec:dataset_summary}
\paragraph{Summary Statistics.} Table~\ref{tab:dataset_statistics} gives the summary statistics for the annotated dataset. \data{} contains a total of \numAnnotatedConstraints{} constraints, applied to \numAnnotatedNorms{} norms, for an average of \constraintsPerConjunction{} constraints per norm. The \framework{} taxonomy broadly captures the kinds of constraints annotators were looking for \taxonomyCoverage{} of the time, and they were able to find their exact constraint value from a pre-populated list in \taxonomyValueCoverage{} cases. For concurrent behavior and attribute constraints, annotators had to input their own values in 59\% and 33\% of cases respectively, followed by 27\% and 9\% of cases for the environment and roles. Overall, this indicates that our GPT-3 prompting method achieved high recall, especially for roles, and least so for behaviors, which is unsurprising, given the almost unbounded space of viable human behavior.

Figure~\ref{fig:dataset_label_proportions} gives the distribution of constraints in \data{}. Constraints are dominated most by the agent's attributes and roles. Age, condition, and characteristic are the most popular attributes, while roles vary. There is an even split between behaviors and the environment. In the environment, there is a notable focus on \textit{time} constraints, and slightly lesser but more even attention towards the remaining subcategories. 

\paragraph{Links to Existing Knowledge.}
\data{'s} \framework{} taxonomy has close links to existing knowledge resources. ConceptNet directly seeded 80 settings in \framework{}. Beyond this, we successfully link over 90\% of taxonomic items from the setting, environment, roles, and attributes directly with concepts in ConceptNet. These taxonomic items cover 93.6\% of all constraint categories and 70.0\% of all constraint values. ConceptNet is further linked to WordNet, DBPedia, Umbel, Cyc, and Wiktionary, so by extension, \data{} can be coupled to these resources.

\section{Experiments: How to use \datae{}}
\begin{table}
\centering
\resizebox{\columnwidth}{!}{%
\begin{tabular}{l|c|c|c|c}
\toprule
  & \multicolumn{4}{c}{\small Norm Classification Results} \\ \midrule
Model & P & R & F & Acc\\ \midrule
ALBERT & 68.0 & 66.6 & 67.1 & 71.0\\
BERT & 72.1 & 70.7 & 71.3 & 74.6\\
RoBERTa & \textbf{73.3} & \textbf{71.4} & \textbf{72.1} & \textbf{75.4}\\ 
\bottomrule
\end{tabular}}
\caption{\textbf{Classification Results} for \{\textit{expected, okay,} or \textit{unexpected}\} show that standard transformer models can learn to make normative inferences with an accuracy that is adequate for expanding \data{}. 
}
\label{tab:clf_results}
\end{table}

\data{} is not designed for any particular narrow task; it is designed as a general-purpose knowledge resource that can ground social reasoning through downstream tasks (compare ATOMIC \citep{sap2019atomic} and ConceptNet \citep{liu2004conceptnet}). Towards this end, \data{} should contain richly organized knowledge that can be learned by neural models and applied for non-monotonic reasoning in new settings. In this way, it should be possible to automatically expand the \data{} resource. The knowledge contained here should also be applicable across a range of social reasoning tasks. Thus our experiments aim to demonstrate two things: (\S\ref{subsec:knowledge_completion}) that we can automatically expand \data{} using neural methods, and (\S\ref{subsec:transfer_learning}) that \data{} is a useful resource with relevant knowledge for downstream applications. For all experiments in the following subsections, we use an 80\%-10\%-10\% train-dev-test split in which \texttt{\small <setting, behavior>} tuples in one set are never seen in another.

\subsection{Automatic Knowledge Completion}
\label{subsec:knowledge_completion}

How can we expand \data{}? We considered two different methods of knowledge bank completion \citep{weston2013connecting,craven2000learning}, which rely on different assumptions. Results from both methods indicate that \data{} is rich enough to support its own automatic expansion. Classification is the simpler case, where we assume a closed world \citep{bordes2013translating,lin2015learning,lin2015modeling,socher2013reasoning}, while generation assumes an open world \citep{shi2018open} with a modifiable set of constraints. 

\paragraph{Classification.} 
Here, our known constraints and behaviors (\S\ref{sec:scene_framework}) will remain fixed \citep{shi2018open}, but we can discover new relationships by classifying unseen behavior and constraint combinations as \textit{expected, okay,} or \textit{unexpected}. The advantage of this approach is that it is straightforward, and the disadvantage is that evaluating classifiers over the power set of the entire constraint space would be intractable; thus more efficient search methods will be needed.

For all norm classification tasks, we fine-tune three popular transformer models: BERT-base-uncased \citep{devlin2018bert}, RoBERTa-base \citep{liu2019roberta}, and ALBERT-base-v2 \citep{lan2019albert}, with hyperparameters in Appendix~\ref{appdx:models_hyperparameters}. Results in Table~\ref{tab:clf_results} appear promising. Given the scope and scale of \data{}, models are capable of learning non-monotonic inferences, achieving F1 scores as high as 72.1\% on the test set. This shows that classification is a reliable method for \data{} knowledge expansion.

\begin{table*}[h!]
\resizebox{\textwidth}{!}{%
\begin{tabular}{lc|cc|ccc|cccc}\toprule
 Model & Decoding & ROUGE-L & BLEU & Avg. \# Constr. & Tax. Constr. & Pre-pop. Constr. & Sensible$_S$ & Correct$_S$ & Norm$_C$ & Relevant$_C$  \\ \midrule
\multirow{3}{*}{BART} & greedy & 75.7 & 23.1 & 1.62 & 93.8 & 40.7
&\best{\textbf{100.0}} & \best{46.0} & \best{\textbf{94.3}} & 94.3\\
 & beam & \best{\textbf{77.1}} & \best{{23.5}} & 1.70 & 98.8 & 48.2
& 98.0 & 36.0 & 92.5 & 94.9 \\
 & p=0.9 & 75.8 & 23.1 & 1.62 & 87.7 & 42.0
& \best{\textbf{100.0}} & \best{46.0} & 91.9 & \best{\textbf{97.0}}\\ \hline

\multirow{3}{*}{GPT-2} & greedy & 57.4 & 12.5 & 1.78 & 70.8 & 36.0
& 86.0 & 34.0 & 86.1 & 89.6\\
 & beam & 70.0 & 16.5 & 1.26 & 95.2 & 47.6
& 94.0 & 34.0 & 88.3 & 94.3 \\
 & p=0.9 & 60.2 & 13.2 & 1.46 & 90.4 & 43.8
& 84.0 & 34.0 & 93.0 & 94.0 \\ \hline

\multirow{3}{*}{T-5} & greedy & 45.0 & 10.3 & 2.70 & 67.2 & 28.4
& 60.0 & 22.0 & 78.8 & 78.9 \\
 & beam & 68.3 & 13.7 & 1.02 & 100.0 & 64.7
& 86.0 & 40.0 & 89.5 & 96.3\\
 & p=0.9 & 48.4 & 10.9 & 2.26 & 72.7 & 30.0
& 76.0 & 42.0 & 83.0 & 79.1 \\ \midrule 
 \multicolumn{2}{l}{GPT-3 davinci-002} & 44.2 & 23.2 & 5.31 & 85.4 & 33.0 & 87.3 & 49.0 & 90.5 & 83.5\\
 \multicolumn{2}{l}{GPT-3 davinci-003} & 51.7 & \textbf{28.6} & 2.34 & 84.6 & 33.2 & 95.0 & \textbf{61.1} & 91.8 & 87.8\\
\hline \midrule
 Gold \data{} && - & - & 2.68 & 93.6 & 70.0 & 82.5 & 55.0 & 72.9 & 81.7 \\ 
\bottomrule 
\end{tabular}
}
\caption{{\textbf{Constraint generation results.} (\textit{Left}) Automatic evaluation suggests that BART has the advantage over other generative models. (\textit{Middle}) Generated constraints fall into the \framework{} taxonomy 67.2-100\% [Tax. Constr.] and use a pre-populated constraint 30-64\% of the time [Pre-pop. Constr.], depending on the decoding strategy.
(\textit{Right}) Human eval shows encouraging results: a \data{}-trained BART can generate sensible, correct, normative, and relevant constraints for use in automatically expanding \data{}. Here, the best fine-tuned model results are \best{highlighted}, while the best overall model results are \textbf{bolded}.}}
\label{tab:gen}
\end{table*}

\paragraph{Generation.} The model is trained with a forward language modeling objective over the string $\boldsymbol{g}$: 

{\small
\begin{align*}
    \boldsymbol{g} &= \{\texttt{[SETTING]}, s_1, s_2, ..., s_n, \\
    & \phantom{abc} \texttt{[BEHAVIOR]}, b_1, b_2, ..., b_m, \\
    & \phantom{abc} \texttt{[NORM]}, \texttt{label},\\
    & \phantom{abc} \texttt{[CONSTRAINTS]} c^1_{1}, c^1_{2}, ..., c^1_{\ell_1}\\
    & \phantom{abc} \texttt{[AND]} c^2_{1}, c^2_{2}, ..., c^2_{\ell_2}...\\
    & \phantom{abc} \texttt{[AND]} c^k_{1}, c^k_{2}, ..., c^k_{\ell_k}\texttt{<EOS>}\}
\end{align*}
}
At inference time, the model generates the list of constraints $c^1, ..., c^k$ that will make the norm \texttt{label} true as it is conditioned on the setting $s$ and behavior $b$. For this purpose, it is sufficient to use BART \cite{lewis2020bart}, GPT-2 \cite{radford2019language}, and T5 \cite{raffel2020exploring}, three powerful language models used widely for generative inference. 
We also prompted GPT-3 davinci-002 and davinci-003 in a few-shot manner via the OpenAI API (see the prompts in Appendix~\ref{appdx:models_hyperparameters}).

Evaluation comes from both automatic and human metrics. Humans evaluate 300 \texttt{\small <setting, behavior>} data points for each of the 11 Model $\times$ Decoding combinations, plus gold standard examples from \data{}. For constraints, they provide us the \textbf{\% Norm$_C$} (\textit{proportion that helps represent a human rule or expectation for behavior}) and \textbf{\% Relevant$_C$} (\textit{proportion that relates to the norm without redundancy or tautology}). They also give us situation\footnote{Situations are defined as the intersection of constraints} evals: are they \textbf{\% Correct$_S$} (\textit{they produce an accurate norm label}) and mutually \textbf{\% Sensible$_S$} (\textit{all constraints can be true at the same time}).

Table~\ref{tab:gen} gives the generation results for the three fine-tuned models: BART, GPT-2, and T-5. According to human judgment, all models produce text that successfully constrains human expectations for behavior (\textit{Norm$_C$} $\sim$90\%). BART + nucleus sampling ($p=0.9$) gives the most \textit{Sensible$_S$} (100\%) and \textit{Correct$_S$} situations (46\%) with the most \textit{Relevant$_C$} constraints (97\%). This is clearly a challenging task: situations are deemed correct only 46\% of the time. Yet they closely approach the scores of human gold-standard data (55\%). Notably, generated constraints are highly relevant to the norm label and entirely mutually-sensible. Given the challenging nature of the task, the results are quite encouraging, suggesting that \data{} can facilitate its own expansion via natural language generation.

\paragraph{Prompting} Results in Table~\ref{tab:gen} show that few-shot GPT-3 models fail to match our best performing BART model's ability to generate \textit{Sensible} situations (95 vs.\ 100\%) with high \textit{\%Norm} (91.8 vs 94.3\%) and \textit{\%Relevant} constraints (87.8 vs 97\%). Still, annotators are more likely to find GPT-3 output to be \textit{Correct} overall (61.1 vs 46\%). Automatic metrics show that GPT-3 achieves higher precision (28.6 vs 23.1 BLEU) at the expense of recall (51.7 vs 77.1 ROUGE-L), suggesting that GPT-3's generations, while often correct, may be more prototypical. Qualitative analysis confirms this.

Sometimes the conventional answer leads GPT-3 astray, as when it uses a series of faulty lexical associations to explain that `drinking milk' is \textit{unexpected} on an `athletic field' for individuals who are not the coach and for those whose behavior is not `hydrate' while the temperature is `warm.' BART, however, correctly discerns that it's unexpected for athletes whose behavior is `playing sports.' In general, GPT-3 appears more likely to underspecify the situation. For example, GPT-3 responds that it's \textit{expected} for a homeowner to `leave the gate open' in the `backyard,' and BART agrees, but BART further specifies that the owner might be `working outside' to justify the \textit{expectation}.

Both quantitative and qualitative analyses indicate that prompting methods can certainly complement, but may not fully replace, fine-tuned generation approaches to \data{} expansion. A mixed approach may be most desirable due to coverage and correctness, while generation errors may be fixed using self-correction via classification (above) or further prompting \citep{fung2022normsage}.

Finally, the middle pane of Table~\ref{tab:gen} shows the proportion of generated constraints that fall into our taxonomy (Tax. Constr.) and the proportion contained in \data{} (Pre-pop. Constr.). The former shows that our taxonomy broadly captures the relevant axes (80-90\% of our best models' generations are taxonomic). The latter shows that between one third and one half of generations `link' prior constraints to new situations; the rest of generated constraints are brand new.

\subsection{Transfer Learning for Downstream Tasks}
\label{subsec:transfer_learning}

Finally, we conduct transfer learning experiments to demonstrate the utility of the data for downstream applications, further indicating the scope and power of \data{} as a general-purpose resource for social reasoning. 
Concretely, we follow the sequential training paradigm \citep{pratt1991direct}, which has proven better than multitask training and fine-tuning on a broad range of commonsense tasks \citep{lourie2021unicorn}. Specifically, we initialize a RoBERTa model with weights from our best-performing norm classifier from Section~\ref{subsec:knowledge_completion} and fine-tune on the target set for 7 epochs.

We evaluate on two specifically \textit{moral} reasoning tasks, Anecdotes and Dilemmas, both from the \textsc{Scruples} benchmark \citep{lourie2021scruples}. We also consider two multiple-choice commonsense QA datasets. Social IQa \citep{sap2019social} is designed to test social intelligence (e.g., inferring motivations, emotional reactions), while CosmosQA \citep{huang2019cosmos} tests cause and effect and counterfactual reasoning in everyday situations.

All results in Table~\ref{tab:transfer_clf_results} are averaged over five separate train-test runs, and significance is given by the paired bootstrap test. \data{'s} utility is seen by comparing the accuracy of models with transfer learning from \data{} against those with task-only fine-tuning (Base Model). Results show that \data{} improves situational moral classification (Anecdotes; +0.4\%) and forced choice binary moral judgments (Dilemmas +6.8\%) with significance. Also consider \data{} utility as compared to transfer learning from either CosmosQA (CQA) or SocialIQa (\textsc{SIQa}). The only task on which transfer from \data{} does not achieve the best performance is on CosmosQA evaluation. Here, we find that transfer from the more structurally related Social IQa task is preferred. We conclude that \data{} is a useful resource for a range of downstream applications in moral, social, and emotional reasoning in context. 

\begin{table}
\centering
\resizebox{1\columnwidth}{!}{%
\begin{tabular}{l|l|l|l|l}
\toprule
Eval & Base & \multicolumn{3}{c}{w/ Transfer Learning from}\\
Task & Model & \textsc{\color[HTML]{CB4335}CQA} & \textsc{\color[HTML]{2E86C1}SIQa} & \textbf{\data{}} \\ \midrule
\texttt{\textsc{anecdotes}} & 68.3 & 68.3\siqa{} & 68.0 & \textbf{68.7}\base{}\siqa{}\\
\texttt{\textsc{dilemmas}} & 64.3 & 67.4 & 70.9\base{} & \textbf{71.1}\base{}\\
\texttt{\textsc{SocialIQa}} & 59.9 & 64.1 & 59.9 & \textbf{64.2}\\
\texttt{\textsc{CosmosQA}} & 59.8 & 59.8 & \textbf{63.5}\base{}\cqa{} & 61.2\\
\bottomrule
\end{tabular}}
\caption{\textbf{Transfer Learning Accuracies} demonstrate the utility of \data{}. By sequential finetuning on \data{}, we improve performance over baseline on all tasks, and transfer performance from \data{} exceeds transfer performance from {\color[HTML]{CB4335}CosmosQA} and from {\color[HTML]{2E86C1}SocialIQa} in three cases. Best performance is \textbf{bolded}. Star\base{} results indicate significant improvements over the Base Model, while \cqa{} marks significance over {\color[HTML]{CB4335}CQA}, and \siqa{} marks significance over \textsc{\color[HTML]{2E86C1}SIQa}.
}
\label{tab:transfer_clf_results}
\end{table}

\section{Conclusion}
Social norms are the foundation of culture and society \citep{mcdonald2015social,hogg2006social}, 
and an understanding of these norms is crucial for assistive and collaborative AI. In this work, we introduced \framework{} a new scheme for hierarchically organizing the seemingly unbounded space of situational contexts that determine social norms. With this framework, we built \data{}, the first social knowledge bank to leverage such contextual information for contrast sets of richly conditioned defeasible social norms. We found that \data{} supports its own automatic expansion via classification, generation, and prompting methods. Finally, we demonstrated the utility of \data{} for situational social reasoning tasks.

\section{Limitations}

At its core, \data{} is a collection of logical operations on unique constraints. 
Consequently, one practical limitation stems from the issue that some situations cannot be reasonably expressed as a set of constraints. While theoretically all logic can be decomposed into AND and OR operations, the logic may be too challenging for an individual to formulate, or the set of constraints themselves might be too large and unwieldy. The latter is problematic, because language models have a finite input token capacity, and for the set of constraints to be digestible, they must fit within that capacity. Relatedly, if the logic to encode constraints become more sophisticated, ensuring that logic is not unnecessarily duplicated will pose a greater challenge. Additionally, certain properties of \data{} like the role and behaviors may be challenging to succinctly describe. Further work will be needed to ascertain how these can be incorporated or to more clearly define situations that are out of scope. 

Due to limitations on time and computational resources, we have not exhaustively evaluated all downstream applications of \data{}, and in future work, we will test additional transfer tasks beyond the moral and social classification tasks considered in this work. Since \data{} is the first to encode non-monotonic situational norms, there was no other available \textit{benchmark} that is directly analogous to ours. Instead, our primary evidence for NormBank’s utility is in Table~\ref{tab:gen}, where human evaluators confirm that models trained on \data{} can reliably learn to make new inferences about non-monotonic situational norms.

Other follow up studies should consider training larger normative reasoning models, and/or engineering better prompts for expanding \data{}. Relatedly, we have no data to speculate about the long-term evolution of real-world norms relative to this resource, nor the rate of decay in the reliability of \data{}. Future work should also expand this resource with perspectives from cultures other than our available annotator pool. The pool was not representative of all cultures and people groups, as we discuss further in the Ethics section.

\section{Ethics}
\label{sec:ethics}

\textbf{Ethical Assumptions.} First, to set proper boundaries on this resource and the tasks it can facilitate, we will outline the ethical assumptions of this work and address some potential misconceptions. We want to stress that \data{} represents a collection of situational norms that we do not treat as prescriptive, but rather descriptive. Unlike prior \textit{moral / ethics} datasets \citep{ziems2022mic, emelin2021moral,lourie2021scruples,forbes2020social,sap2020social}, we use the neutral language of \textit{expected, okay,} and \textit{unexpected} behaviors to focus on empirically observed patterns and avoid an over-emphasis on the ethical grey area of what \textit{ought} to be done. Unlike tricky moral dilemmas, the situational social norms of \data{} have an answer that a majority can agree is descriptively observable as the expectation under the respective conditions and/or cultural context. Nevertheless, normative judgments can vary between individuals in different social groups and time periods  \cite{haidt1993affect,shweder1990defense,bicchieri2005grammar,culley2013note,amaya2021new}. \data{} can and should be expanded via automatic or manual methods that can incorporate these axes of variation. Our annotator pool was limited to English-speaking individuals living in the United States in the year 2022. Future expansion efforts could be crowdsourced from other cultures and geographic regions and in future decades.

We reiterate that the norms in \data{} should \emph{not} be used for prescriptive advice or personal guidance in any way. Our work intends to unlock future work in the capacity to imbue language models with situational commonsense and enable them to jointly reason with the situational contexts. Language models which ignore situational contexts altogether may be just as hazardous, if not more. 

Finally, there are likely biases towards certain roles and values in \data{}. We have taken steps to mitigate some forms, such as gender bias, by neutralizing constraints (e.g., [PERSON]'s role is `cowboy or cowgirl' and [PERSON]'s role is `ball boy or girl'). Our \framework{} taxonomy, with the standardized structure of its role and attribute constraints, will allow practitioners to further analyze specific axes of prejudice and thus implement targeted mitigation strategies. Specific identity attributes like gender, ethnicity, and religion are represented in 24\% of norms.\footnote{There are 40k norms (out of 169k total norms; 24\%) which cover attributes in the following set: \{`country', `age bracket', `education', `gender', `race or ethnicity', `religion', `sexuality', `social class'\}} Stakeholders can invest a smaller but more concerted effort towards mitigating bias in these constraints. We encourage stakeholders to give auditing control over a given norm to those who are affected by it. Previous norm-datasets encode norms in free-text annotations which lack a hierarchical taxonomy of contexts, but our taxonomy can be used to interpret, diagnose, and mitigate prejudice, and to return power to those affected by these prejudices.

\paragraph{Risks in deployment.} Before starting any annotation, the resources and findings presented in this work were thoroughly reviewed and approved by an internal review board. Prior to being put into production, the method would also need to be re-evaluated when applied to a new domain to ensure reliable performance in order to prevent unintended consequences. To help mitigate risks in deployment from misunderstandings about the ethical assumptions above, we require users of this data to complete a Data Use Agreement. The user will check that they understand the ethical assumptions above: especially that \data{} is not to be taken for advice. Practitioners will also agree not to use \data{} for malicious purposes ``including (but not limited to): mockery, discrimination, and hate speech.'' 

\section*{Acknowledgements}  We are thankful to Julia Kruk, William Held, Albert Lu, Camille Harris, and the anonymous ACL reviewers for their helpful feedback. Caleb Ziems is supported by the NSF Graduate Research Fellowship under Grant No. DGE-2039655.

\bibliography{ref}
\bibliographystyle{acl_natbib}
\clearpage
\appendix
\section{Models \& Hyperparameters}
\label{appdx:models_hyperparameters}
\paragraph{Classification.} We use the base versions of BERT (\citeauthor{devlin2018bert}; 768-hidden, 12-heads, 110M parameters), RoBERTa (\citeauthor{liu2019roberta}; 768-hidden, 12-heads, 125M parameters), and ALBERT-v2 (\citeauthor{lan2019albert}; 768-hidden, 12-heads, 11M parameters). For each model, we fine-tune using AdamW \citep{adamW} for 7 epochs with a batch size of 16 and a learning rate of $1e-5$. These hyperparameters were chosen by hyperparameter search on the dev set over $\{1e-5, 2e-5, 3e-5, 5e-5\}$ and the number of epochs in $\{1 .. 8\}$, with $\epsilon=1e-8$ and the batch size set to 16.

\paragraph{Generation.} We trained BART-large (406M parameters), GPT-2 (768-hidden, 12-heads, 117M parameters), and T5-small (512-hidden, 8 heads, 60M parameters) for 1 epoch using a batch size of 8 and a learning rate of $3e-5$. We also prompted GPT-3 Davinci-002 and Davinci-003 in a few-shot manner via the OpenAI API, using the following prompt, which contains seven representative datapoints from the \data{} train set:
\begin{lstlisting}
    It is UNEXPECTED to "talk about sex" at an "cafe" when:
+ PERSON's role is 'barista'

It is OKAY to "listen to music" in an "arcade" when all of the following are true:
+ PERSON's role is not 'cook'
+ PERSON's role is not 'maintenance worker'
+ PERSON's behavior is 'wear headphones'

It is EXPECTED to "hang film to dry" at the "darkroom" when all of the following are true:
+ PERSON's role is 'darkroom technician'
+ PERSON's behavior is 'performing routine tasks'

It is OKAY to "smoke a cigar" in a "limousine" when all of the following are true:
+ OTHER's age bracket is not 'gradeschooler or younger'
+ OTHER's attitude towards smoking is not 'bothered by smoke'
+ PERSON's age bracket is 'adult'
+ limousine rules is not 'forbids smoking'

It is EXPECTED to "maintain the store appearance" at the "clothing store" when all of the following are true:
+ PERSON's role is 'store owner'

It is OKAY to "talk to yourself" on an "airplane" when all of the following are true:
+ PERSON's role is 'passenger'
+ PERSON's behavior is 'pray'

It is UNEXPECTED to "grow tomatoes" in a "garden" when all of the following are true:
+ PERSON's behavior is not 'use a greenhouse'
+ temperature is 'freezing'

It is {norm} to "{behavior}" {prep} {det} "{setting}" when all of the following are true:
+
\end{lstlisting}

\section{Additional Details on Constructing \framework{}}

\subsection{The Environment \environment{}}
\label{appdx:environment}

\paragraph{Country of Operation} is seeded with the 195 countries from the \citet{UN} list of member or non-member observer states.

\paragraph{Operational Factors} is a broad category of constraints from environmental psychology \citep{bell2001environmental} involving one's comfort and ease of operation in an environment. Such factors influence descriptive norms. Operational behaviors can be influenced by the degree of sensory stimulation, as well as by {privacy} and {proxemics}, or the local density and organization of persons and objects \citep{russell1982environmental}. These inform the following subcategories: \textbf{attendance} \{\textit{empty, there are people around, crowded}\} \citep{altman1975environment}; \textbf{cleanliness} \{\textit{dirty, clean}\} \citep{vilnai2010effect,cialdini1990focus}; \textbf{noise} \{\textit{quiet, moderate, loud}\} \citep{mathews1975environmental}; \textbf{population density} \{\textit{urban, suburban, rural}\} \citep{scott2007urban}; and \textbf{privacy} \{\textit{private, public}\} \citep{altman1975environment}. 

\vspace{-3pt}
\paragraph{Physical Conditions} in the environment can influence behavior mechanically as well as psychologically. Specifically, the \textbf{lighting} \{\textit{bright, moderate, dim, dark}\}, \textbf{weather} \{\textit{blizzard, clear, cloudy, ...}\}, and \textbf{temperature} \{\textit{freezing, cold, temperate, hot}\} can directly impact visibility, coordination, and the perception of safety \citep{boyce2000perceptions,van2018illuminating}, as well as comfort, confidence, and altruism \citep{cunningham1979weather}.

\vspace{-3pt}
\paragraph{Restrictions} formally limit attendance, participation, and behavior. The environment can be one of \textbf{exclusion} \{\textit{adults only, men only, women only}\}, \textbf{formality} \{\textit{formal, informal}\} or \textbf{religiosity} \{\textit{sacred, secular}\}. These categories are not part of our original theoretical taxonomy but were introduced through annotator feedback (See Section~\ref{sec:annotation}).

\vspace{-3pt}
\paragraph{Special Observances} include cultural observances like \textbf{holidays} \{\textit{Advent, Holi, Lunar New Year}\} as well other \textbf{special events} \{\textit{bat mitzvah, housewarming, quinceañera}\}, which evoke distinct rituals, customs and norms \citep{durkheim1915elementary}. 

\vspace{-3pt}
\paragraph{Time} constraints include \textbf{day of the week}, \textbf{season}, \textbf{time of day}, and \textbf{time period}. Like special observances, much of human activity adheres to a set of temporal constraints and cues \citep{janicik2003talking}.

\subsection{Behaviors \behavior{}}
\label{appdx:behaviors}
Since the average string length of behaviors was greater and thus more prone to error, we applied a suite of programmatic cleaning and filtering techniques, followed by a manual filtering round that reduced the average to 112.6 clean and non-redundant behaviors per setting. The filtering techniques are as follows:

\begin{enumerate}
    \item Remove any conditional form ``if you were...'' as well as any mention of the role or the setting in the behavior itself
    \item Remove elaborations on behaviors ``<behavior> because ... <elaboration>''
    \item Normalize the logical form by removing words that negate behaviors (do not, never, not, forget to, refuse to, fail to, anything, something, anyone, someone, in any way, any)
    \item Remove biased terms like \textit{properly, should, would, try to, be able to, and see someone}.
    \item Remove any bullet points or stray characters
    \item Require that the behavior has an active (not passive) verb in it and that the verb does not have an explicit subject and that there is no dependent clause (as indicated by the marker [\texttt{mark}] dependency).
\end{enumerate}

\section{Annotation Task Details}

\subsection{Qualification Task}
\label{appdx:qualification_test}
To qualify for the HIT, workers were required to pass the following qualifying test, answering at least 5 out of 6 questions correctly.

\begin{enumerate}
    \item \textbf{True or False:} you are allowed to add your own Constraints by typing them directly into the box. [\textit{Answer}: \textbf{True}]
    \item \textbf{True or False:} ``carry a gun'' can be an ``Expected'' behavior on an Airplane. [\textit{Answer}: \textbf{True}]
    \item \textbf{True or False:} ``read a book'' is an ``Expected'' behavior for a passenger on an Airplane. [\textit{Answer}: \textbf{False}]
    \item \textbf{True or False:} it is possible for a BEHAVIOR to be both ``Expected'' and ``Okay'' under the same Constraints. [\textit{Answer}: \textbf{False}]
    \item Let's say you are adding some Constraints for when ``eating shrimp'' is ``Unexpected'' in the SETTING: restaurant. You know that shellfish are disallowed in both Hinduism and Judaism, as well as by the vegan and vegetarian diets. You are thinking of adding these in the following Constraint table. Is this correct? [\textit{Answer}: \textbf{Incorrect}]
    \begin{quote}
        (PERSON's religion is Judaism) AND (PERSON's religion is Hinduism) AND (PERSON's diet is vegan) AND (PERSON's diet is vegetarian)
    \end{quote}
    \item Let's say you are adding some Constraints for when "eating shrimp" is "Okay" in the SETTING: restaurant. You know that shellfish are disallowed in both Hinduism and Judaism, as well as by the vegan and vegetarian diets. You are thinking of adding these in the following Constraint table. Is this correct? [\textit{Answer}: \textbf{Correct}]
    \begin{quote}
        (PERSON's religion is NOT Judaism) AND (PERSON's religion is NOT Hinduism) AND (PERSON's diet is NOT vegan) AND (PERSON's diet is NOT vegetarian)
    \end{quote}
\end{enumerate}

\subsection{HIT Interface}
\label{appdx:hit_interface}

\begin{figure*}
    \centering
    \includegraphics[width=0.54\textwidth]{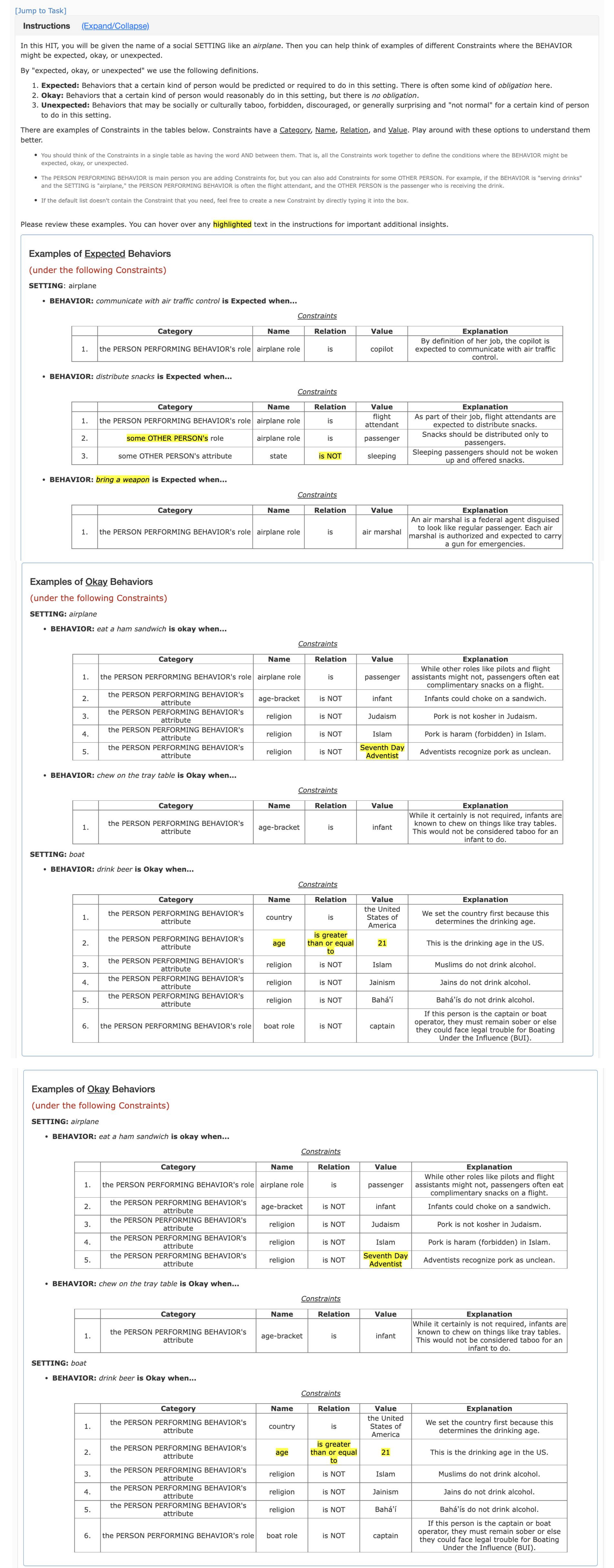}
    \caption{HIT Instructions.}
    \label{fig:hit_instructions}
\end{figure*}

\begin{figure*}
    \centering
    \includegraphics[width=0.47\textwidth]{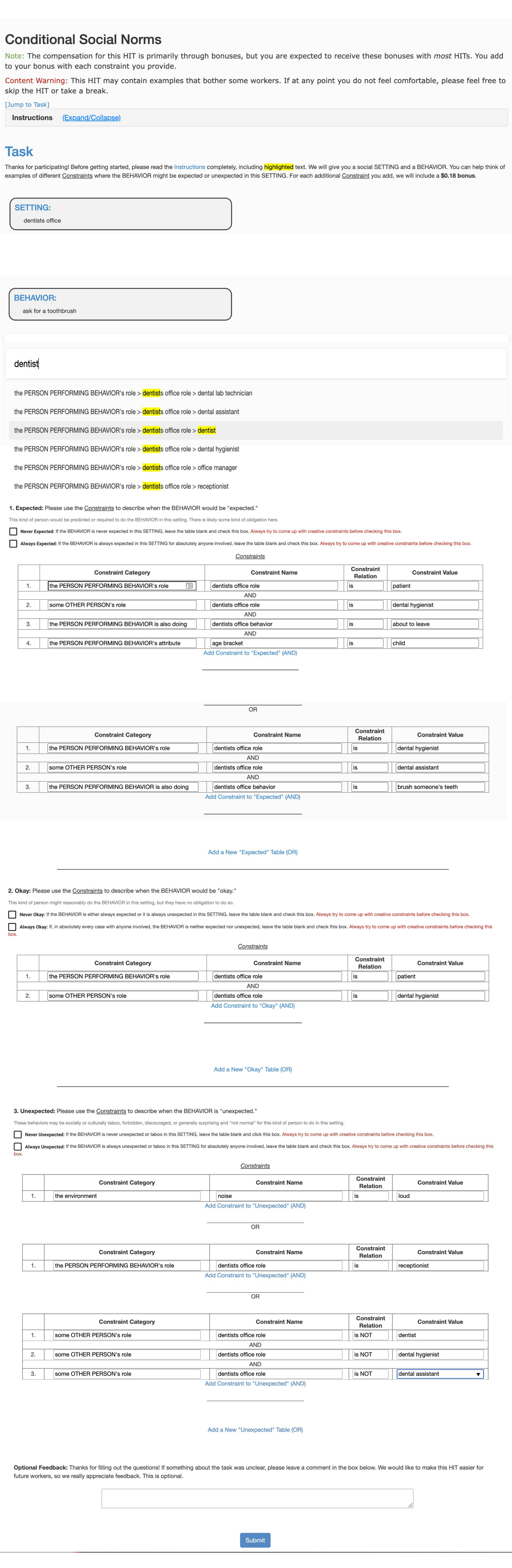}
    \caption{HIT Interface}
    \label{fig:hit_interface}
\end{figure*}

For each HIT, the annotator is presented with a setting $s \in \mathcal{S}$ and a behavior $b \in \mathcal{B}$ that we generated for the given $s$. The annotator helps us describe when this behavior would be \textit{expected}; then describes when it is merely \textit{okay}; and finally \textit{unexpected}. Annotators describe each norm with the conjunction and disjunction of \framework{} constraints. The annotator appends each constraint to its conjunction as a 4-tuple consisting of a (1) \textit{category}, (2) \textit{name}, (3) \textit{relation}, and (4) \textit{value}. These are shown with examples in the HIT Instructions (\ref{fig:hit_instructions}) and HIT Interface (Figure~\ref{fig:hit_interface}) screenshots. The \textit{category} helps annotators search for constraints and organize their thoughts. The \textit{category} is a high-level designation of where the constraint is organized: according to the \textit{environment}, \textit{role}, \textit{attribute}, or \textit{behavior}. The \textit{name, relation} and \textit{value} constitute a standard semantic triple. The \textit{name} designates the subject of the constraint, and it a specification of the \textit{category}, like the ``temperature of the environment.'' The \textit{relation} is a logical type that includes equality and inequality. The \textit{value} designates the predicate of the constraint (e.g., ``{freezing}'').

Annotators can build constraint 4-tuples from drop-down menus that enumerate our hierarchical taxonomy (Section~\ref{sec:scene_framework}). Annotators can also freely edit the above fields and contribute novel constraints. Finally, annotators compose constraints into disjunctive normal form (DNF), the OR of ANDs, to describe when behaviors are \textit{expected, okay,} or \textit{unexpected} in a given setting.

\end{document}